%% file: main.tex
\documentclass[letterpaper, 10 pt, conference]{ieeeconf}  

\IEEEoverridecommandlockouts                              

\usepackage[dvipdfmx]{graphicx} 
\usepackage{multirow}
\usepackage{arydshln}

\title{\LARGE \bf
Sensorimotor Attention and Language-based Regressions in Shared Latent Variables
for Integrating Robot Motion Learning and LLM
}

\author{
Kanata Suzuki$^{1,2}$ 
and Tetsuya Ogata$^{1,3}$
\thanks{
All authors are affiliated with Faculty of Science and Engineering, Waseda University, Tokyo 169-8050, Japan. 
$^{1,2}$Kanata Suzuki is also at Artificial Intelligence Laboratories, Fujitsu Limited., Kanagawa 211-8588, Japan. 
$^{1,3}$Tetsuya Ogata is also at the National Institute of Advanced Industrial Science and Technology, Tokyo 100-8921, Japan. 
E-mail:{\tt\small suzuki.kanata@fujitsu.com}
}}

\begin{document}

\maketitle
\thispagestyle{empty}
\pagestyle{empty}


\input{sections/0_abstract}

\section{INTRODUCTION}
\label{sec:introduction}
\input{sections/1_introduction.tex}

\section{RELATED WORK}
\label{sec:related_work}
\input{sections/2_related_work.tex}

\section{PROPOSED METHOD}
\label{sec:method}

\input{sections/3_method}

\section{EXPERIMENTS}
\label{sec:experiment}

\input{sections/4_experiments}

\section{RESULTS AND DISCUSSION}
\label{sec:result}

\input{sections/5_results}

\section{CONCLUSION}
\label{sec:conclusion}

\input{sections/6_conclusion.tex}




\section*{ACKNOWLEDGMENT}
This work was supported by JST Moonshot R\&D Grant Number JPMJMS2031 and JSPS Grant-in-Aid for Early-Career Scientists (Grant Number: 24K20877), Japan.

\bibliographystyle{IEEEtran}
\bibliography{main}

\end{document}

%% file: sections/0_abstract.tex
\begin{abstract}


In recent years, studies have been actively conducted on combining large language models (LLM) and robotics; however, most have not considered end-to-end feedback in the robot-motion generation phase. 
The prediction of deep neural networks must contain errors, it is required to update the trained model to correspond to the real environment to generate robot motion adaptively. 
This study proposes an integration method that connects the robot-motion learning model and LLM using shared latent variables. 
When generating robot motion, the proposed method updates shared parameters based on prediction errors from both sensorimotor attention points and task language instructions given to the robot. 
This allows the model to search for latent parameters appropriate for the robot task efficiently. 
Through simulator experiments on multiple robot tasks, we demonstrated the effectiveness of our proposed method from two perspectives: position generalization and language instruction generalization abilities.

\end{abstract}

%% file: sections/1_introduction.tex

The application of large language models (LLMs) to robotics is actively being explored in many studies~\cite{Shridhar2021a}\cite{Liang2022a}\cite{Brohan2022a}\cite{Driess2023a}. 
Many practical systems designed the LLM and robot-motion control unit independently and demonstrated high generalization ability in zero-shot inference. 
However, in such cascade-type configurations, it is difficult to incorporate end-to-end feedback for updating the model from the robot at the raw sensor level. 
The predictions of deep neural networks (DNNs) trained with offline data must contain errors, and thus, to generate robot motion adaptively, it is necessary to update the model to correspond to the real environment. 
Although there have been studies that used dialogue with collaborators or foundation models to determine the success or failure of robot tasks~\cite{Huang2022b}\cite{Huang2023}\cite{Ren2023a}, they did not consider the feedback necessary for online control at the joint motor level.


On the other hand, there are some methods for building the DNN model from scratch by learning multimodal data including robot motion sequences~\cite{Brohan2022a}\cite{Driess2023a}. 
These methods collect paired data of motion and other modalities (language, images, etc.) by operating multiple robots over long periods. 
By fine-tuning additional parameters while fixing some or all of the learned weights of the LLM, the model can connect the modality representation with the real environment. 
Although these are promising efforts to overcome the limitations of LLMs, they do not consider online feedback because of the high cost of data collection and model training.


To summarize, integrated learning methods for LLMs and robot motion need to be able to adaptively predict using a small training cost. 
In the field of natural language processing, past studies changed the behaviors of their models without extensive training by partially fine-tuning the LLM weights~\cite{hu2022lora} or attaching external storage~\cite{lewis2020retrieval}. 
The approach of connecting the motion learning model and language model using shared parameters has also been explored in a variety of fields, including cognitive robotics~\cite{Tani2002a}\cite{Ogata2007a}\cite{Ito2022a}. 
Considering the grounding of language and robot motion~\cite{alter3}, it is important to update the connecting parameters using both language and sensor information.

\begin{figure}[t]
\centering
\includegraphics[width=0.8\columnwidth]{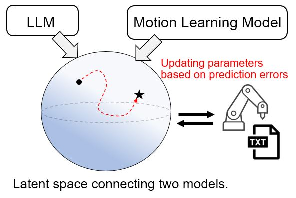}
\caption{
Overview of this study. In the proposed method, latent variables related to robot tasks are updated based on prediction errors for instruction sentences and sensorimotor attention.
}
\label{fig:fig1}
\end{figure}

In this study, we propose an integration framework that connects the motion learning model and LLM using shared latent variables. 
We use a Spatial Attention Transformer RNN (SATrRNN), which predicts attention points for robot task learning, and a Receptance Weighted Key Value (RWKV~\cite{Peng2023a}), which is an LLM with a linear attention mechanism. 
When generating robot motions, the proposed method updates shared parameters based on prediction errors from both attention points for sensor information and task instruction sentences given to the robot (Fig.~\ref{fig:fig1}). 
This allows the model to efficiently search for parameters appropriate for the target robot task in its latent space. 
We conducted learning experiments for manipulation tasks on a simulator and demonstrated the effectiveness of our proposed method.

\setlength\textfloatsep{5pt}

%% file: sections/2_related_work.tex
\setlength\textfloatsep{5pt}
\begin{figure*}[t]
\centering
\includegraphics[width=2.0\columnwidth]{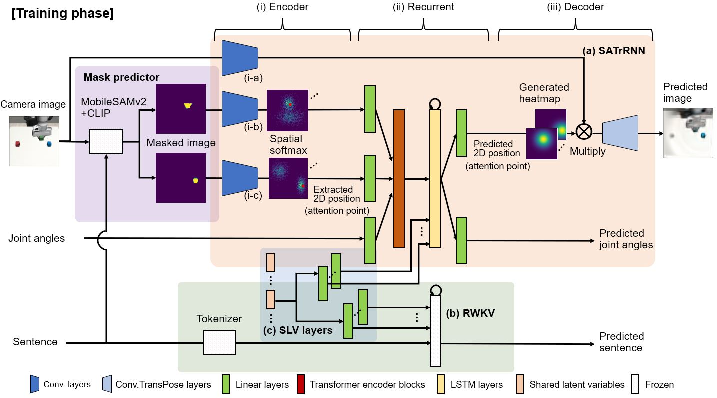}
\caption{
Overview of the proposed method, consisting of three modules: SATrRNN with mask predictor, RWKV, and shared latent variables.
}
\label{fig:fig2}
\end{figure*}


The generalization ability of foundation models has greatly increased the possibility of real-world applications in robotics. 
In particular, task planning~\cite{Liang2022a}\cite{Ahn2022a}\cite{Huang20222a}\cite{Vemprala2023} and object recognition methods~\cite{Zhou2022a}\cite{Kirillov2023a}\cite{Cheng2024a} have shown higher accuracies than those of conventional methods for connecting natural language and our daily environment. 
However, the prediction results of DNN models trained with offline data may be inconsistent with the behavior of a real robot. 
To address this problem, Brohan et al. performed language-conditioned robot tasks in a real environment by connecting LLM's task planning with a motion learning model~\cite{Brohan2022a}. 
They developed the system for determining the next action by linking a model that predicts the probability of successfully executing a skill with an LLM that evaluates the probability of selecting each skill. 
However, because the skills are predefined, the model tends to lack versatility in terms of the tasks that it can accomplish and requires a large amount of data for its training.


On the other hand, approaches to modifying prediction results have also been actively studied~\cite{Huang2022b}\cite{Huang2023}\cite{Ren2023a}\cite{Skreta2024a}\cite{Hori2024a}. 
Huang et al. incorporated closed-loop verbal feedback based on action success signals into task planning and performed robot tasks with high-level commands~\cite{Huang2022b}. 
Ren et al. proposed a method that presents actions with confidence ratings during task planning using an LLM and poses questions to collaborators when uncertainty is high~\cite{Ren2023a}. 
Hori et al. attempted to use an LLM to analyze the uncertainty of task planning and improve long-horizon planning results through a question-and-answer approach~\cite{Hori2024a}. 
However, these past studies did not consider sending feedback to the LLMs at the sensor information level, which is important for online robot-motion generation.


Studies on connecting language and robot-motion learning have been conducted since before the LLM era~\cite{jang2021bc}\cite{Toyoda2021a}\cite{Toyoda2022a}\cite{Jiang2023a}\cite{Ito2022a}. 
Particularly in DNN-based methods, there has been a study that integrated language and motion representations by bringing them closer together in latent space~\cite{jang2021bc}. 
It has also been reported that the bidirectional transformation of modalities can be accomplished using a common latent space~\cite{Toyoda2021a}\cite{Toyoda2022a}\cite{Jiang2023a}. 
In this way, learning constraints regarding multiple modalities on the latent space are effective for interactive robot control.


The approach that we propose in this study is to perform error regressions at the motion generation phase for latent parameters that connect an LLM and a motion learning model. 
This approach is based on the concept of predictive coding~\cite{Friston2006jpp}, which controls model behavior based on minimizing prediction errors, and we apply it to robot learning~\cite{Suzuki2023}. 
By using language and attention information based on robot motion learning as a prediction error, it is expected that parameters are updated to reflect more important grounding information in the robot task. 
To the best of our knowledge, the integration of an LLM and the robot learning model using this approach has not been performed in past studies and is an important contribution of this study.

%% file: sections/3_method.tex

The proposed model consists of a spatial attention transformer RNN (SATrRNN) that predicts time sequences of sensorimotor information; and RWKV~\cite{Peng2023a}, a lightweight LLM. 
This model expands a past study on language-motion integrated learning~\cite{Ito2022a}. 
We prepare common latent variables that are connected to the hidden layer of the two models and simultaneously optimize the entire model during the training phase. 
An overview of the proposed model is shown in Fig.~\ref{fig:fig2}. 
Table I shows a specific configuration of the model, including its layer types and parameters.

\begin{figure}[t]
\centering
\includegraphics[width=\columnwidth]{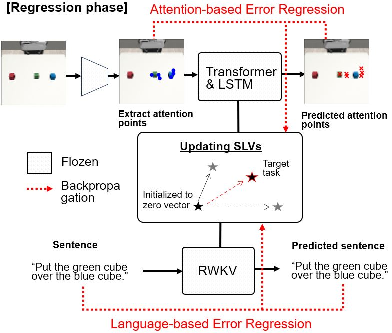}
\caption{
Overview of the proposed error regression method. 
The SLV is optimized from the reconstruction error for language instruction and MSE between extracted attention points (blue circle marks) and predicted attention points (red cross marks).
}
\label{fig:fig3}
\end{figure}

The key idea of the proposed method is to construct a common latent space by learning together with the motion learning model while keeping the weight parameters of the LLM fixed. 
This makes it possible to update task-appropriate latent variables during the motion generation phase via the backpropagation of raw sensor information and language prediction without extensive learning (Fig.~\ref{fig:fig3}).

\setlength\textfloatsep{5pt}

\subsection{Model Architecture}
\subsubsection{SATrRNN}

The SATrRNN consists of three units: an encoder unit (Fig.~\ref{fig:fig2}-i), a recurrent unit (Fig.~\ref{fig:fig2}-ii), and a decoder unit (Fig.~\ref{fig:fig2}-iii). 
The encoder part has one image encoder (i-a) and two attention point encoders (i-b/c). 
From a 64$\times$64$\times$3-pixel camera image, the image encoder extracts a feature map, whereas each attention point encoder extracts three position coordinates using spatial softmax~\cite{finn2016deep}. 
Because two attention point encoders are used, a total of 6 coordinates are extracted. 
In this study, to enable robust feature extraction for the background or similar objects, we used a mask image predicted by MobileSAMv2~\cite{Zhang2023a} as input to the attention point encoders. 
We used CLIP~\cite{Radford2021a} to predict the degree of similarity between the instruction sentence and each mask region, and adopted the top two that were above the threshold. 
In cases wherein only one region is detected, the same one is input into the attention point encoders. 
It is expected that through the use of multiple encoders to extract coordinates, attention points will be predicted for each object.


The extracted 6 position coordinates and joint angles are used as input tokens for a 4-layer transformer encoder to learn the relationships between modalities. 
The self-attention layers of the transformer encoder predict feature vectors useful for task learning from redundantly extracted position coordinates. 
Then, a 3-layer LSTM network predicts the next state of the sensorimotor sequence. 
The output of the LSTM is converted into joint angles and position coordinates for the next step through the linear layer. 
The position information predicted by the LSTM is converted into heatmaps, and the image decoder outputs the next step image based on the feature map and heatmap. 
Image encoders and decoders extract information about the colors and shapes of objects by reconstructing images. 
This allows the SATrRNN to generalize object position information, learn their relationship with joint angles, and generate appropriate motions.

\begin{table}[t]
    \centering
    \begin{tabular}{cclll}
        \multicolumn{5}{c}{TABLE I: Structure of Networks} \\
        \hline
        & & Layer type & Parameter & Output shape \\
        \hline \hline
        & \multirow{3}{*}{(i-a)} 
          & Conv2D 1        & k=3, s=1, ch=18 & 62$\times$62$\times$18 \\ 
        & & Conv2D 2        & k=3, s=1, ch=36 & 60$\times$60$\times$36 \\ 
        & & Conv2D 3        & k=3, s=1, ch=6  & 58$\times$58$\times$6 \\ 
        \hdashline
        & \multirow{4}{*}{(i-b/c)} 
          & Conv2D 4, 7     & k=3, s=1, ch=9  & 62$\times$62$\times$9 \\ 
        & & Conv2D 5, 8     & k=3, s=1, ch=18 & 60$\times$60$\times$18 \\ 
        & & Conv2D 6, 9     & k=3, s=1, ch=3  & 58$\times$58$\times$3 \\ 
        & & Spatial Softmax & -              & 3$\times$2 \\ 
        \hdashline
        \multirow{7}{*}{(a)} & \multirow{7}{*}{(ii)} 
          & Linears 1        & layer=1, hid=20  & 20 \\ 
        & & Linears 2        & layer=1, hid=20  & 20 \\ 
        & & Transformer Enc. & layer=4, head=4  & 20$\times$7 \\ 
        & & LSTMs            & layer=3, hid=100 & 100 \\ 
        & & Linears 3        & layer=1, hid=12  & 12 \\ 
        & & Linears 4        & layer=1, hid=8   & 8 \\ 
        \hdashline
        & \multirow{5}{*}{(iii)} 
          & Generate heatmap  & -              & 58$\times$58$\times$6 \\ 
        & & Multiply          & -              & 58$\times$58$\times$6 \\ 
        & & Conv2DTrans. 1    & k=3, s=1, ch=18 & 60$\times$60$\times$18 \\ 
        & & Conv2DTrans. 2    & k=3, s=1, ch=36 & 62$\times$62$\times$36 \\ 
        & & Conv2DTrans. 3    & k=3, s=1, ch=3  & 64$\times$64$\times$3 \\ 
        \hline
        \multirow{2}{*}{(b)} 
        & & RWKV block & layer=12, hid=768 & 768 \\ 
        & & RWKV head  & hid=50277         & 50277 \\ 
        \hline
        \multirow{3}{*}{(c)} 
        & & Variable      & dim=5   & 5 \\ 
        & & Linears 5--7  & layer=2, hid=100 & 100 \\ 
        & & Linears 8--19 & layer=3, hid=768 & 768 \\ 
        \hline
    \end{tabular}
    \begin{flushleft}
        * All layers have leaky ReLU activation. \\
        *** Linear and LSTM layers have layer normalization. \\
        **** k: kernel, s: stride, ch: channel dim, layer: layer num, hid: hidden dim. \\
    \end{flushleft}
\end{table}

\subsubsection{RWKV}

We use RWKV as a unit that accepts language instructions from collaborators and sends them to robots (Fig.~\ref{fig:fig2}b). 
The RWKV has a structure in which time-mixing blocks, which extend the linear attention mechanism, and channel-mixing blocks are stacked alternately. 
It is possible to parallelize the training process and maintain the cost in terms of amount of calculation when performing inference. 
The RWKV consists of 12 blocks, each of which has a hidden state dimension of 768. 
We adopted a particularly lightweight model because memory efficiency is important for robot motion generation. 
The instruction sentences are converted into tokens by a tokenizer, and the RWKV makes predictions to restore the input. 
The weight parameters of the RWKV are initialized with trained weights and are not updated during both the training and inference phases. 
Therefore, the prediction of the RWKV is modified only by shared parameters, which are described in the next subsection.

\subsubsection{Shared Latent Variables}

We incorporate parametric bias~\cite{Tani2002a} as the shared latent variable (SLV) into the model to obtain integration representations of language and motion (Fig.~\ref{fig:fig2}c). 
The SLV is a 5-dimensional vector and is initialized to zero. Unlike other input values, the SLV does not depend on the time series and is always a constant value. 
The SATrRNN and RWKV are independent models, and each model is connected through multiple linear layers such that the SLV is shared. 
The SLVs are prepared for each motion sequence in the dataset and learn the correspondence between language and motion as a variable by updating individual parameters for each training sequence. 
Because the SATrRNN and RWKV use common weights for all sequences in the training phase, the SLV strongly maps the relationship between language and motions. 
Therefore, the SATrRNN can predict appropriate task motions based on bottom-up recognition of object position coordinates using the SLV as input to LSTM layers.

\subsection{Training and Regression Phases}
During the training phase, the models are optimized using an imitation learning scheme. 
The SATrRNN extracts attention points $pt^{enc}_{t}$ from camera image $i_t$ using the image encoders. 
Then, using $pt^{enc}_{t}$ and joint angle $j_t$ as input, the LSTM predicts the attention points $pt^{dec}_{t+1}$ and $j_{t+1}$ for the next step. 
The image decoder receives $pt^{dec}_{t+1}$ as input and reconstructs the next step image $i_{t+1}$. The RWKV is configured to reconstruct the tokenized input sentences $s$. 
The overall loss function $L_{train}$ is expressed as follows.
\begin{eqnarray}
    L_{train} = \alpha L_{ja} + \beta (L_{img} + L_{pt}) + \gamma L_{dsc} \\
    L_{ja} = \frac{1}{T_s-1} \sum_{t=1}^{T_s-1} \left[ (j_{t+1}-\hat{j}_{t+1})^2 \right] \\
    L_{img} = \frac{1}{T_s-1} \sum_{t=1}^{T_s-1} \left[ (i_{t+1}-\hat{i}_{t+1})^2 \right] \\
    L_{pt} = \frac{1}{T_s-1} \sum_{t=1}^{T_s-1} \left[ (p^{enc}_{t+1}-p^{dec}_{t+1})^2 \right]\\
    L_{dsc} = \frac{1}{T_d-1} \sum_{t=1}^{T_d-1} \left[ \sum_{w}^{W} \hat{s}_{t+1,w} \log s_{t+1,w} \right]
\end{eqnarray}
where $L_{ja}$ represents the mean square error (MSE) of the joint angle; $L_{img}$ the MSE of the predicted image; $L_{pt}$ the MSE of the attention points $p_t = (x_t, y_t)$; $L_{dsc}$ the cross-entropy loss of sentence prediction by RWKV; $T_s$ the motion sequence length; $T_d$ the sentence length; $\hat{\bullet}$ the target value obtained from the training dataset; and $W$ the vocabulary size of the RWKV.
$\alpha$, $\beta$, and $\gamma$ represent the loss contribution, of which only $\alpha$ was set to 1.0, whereas $\beta$ and $\gamma$ were set to 0.1. 
RAdam~\cite{Liu2019a} was used as the optimization method, and the model was trained for a total of 5000 epochs.

In robot-motion generation, the SLV value is initialized to zero and a regression phase is provided to search for an appropriate SLV value before starting the task (Fig.~\ref{fig:fig3}). 
The regression loss $L_{ER}$ uses loss terms related to the language and attention points. 
Because the correspondence between language and motion is embedded in the SLV, through minimization of $L_{ER}$ and optimization for the SLV, the behavior of the model is changed to predict the object position and the corresponding task motion appropriately. 
The weight parameters other than the SLV are fixed, and the SLV is updated to minimize the following loss function:
\begin{eqnarray}
    L_{ER} = \sum_{t=1}^{T=10} \left[ (p^{enc}_0-p^{dec}_1)^2 + L_{dsc} \right]
\end{eqnarray}
Here, $L_{pt}$ targets only the initial environment state. 
We use RAdam as the optimization method and update the parameters for 10 iterations. 
Using the SLV values updated through the regression phase, the model predicts the next state from the current joint angles and camera images. 
The target posture of the robot is defined based on the predicted joint angles, and the joints are controlled to generate motion online.

\setlength\textfloatsep{5pt}
\begin{figure}[t]
\centering
\includegraphics[width=\columnwidth]{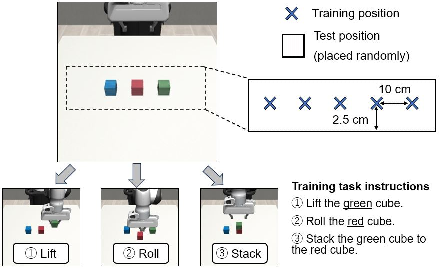}
\caption{
Robot task setup in our experiments.
}
\label{fig:fig4}
\end{figure}

%% file: sections/4_experiments.tex
\setlength\textfloatsep{5pt}
\begin{figure}[t]
\centering
\includegraphics[width=1.0\columnwidth]{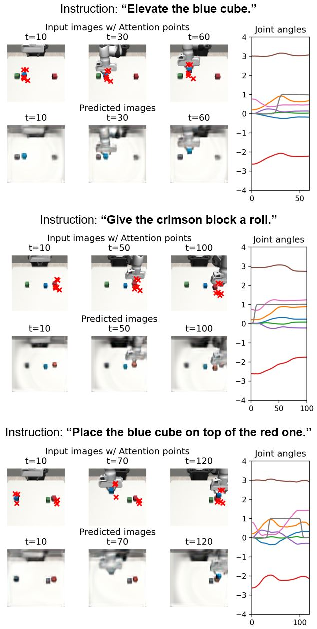}
\caption{
Examples of generated Lift, Roll, and Stack task sequences in case 2 (test position).
}
\label{fig:fig5}
\end{figure}

\begin{figure*}[t]
\centering
\includegraphics[width=2.0\columnwidth]{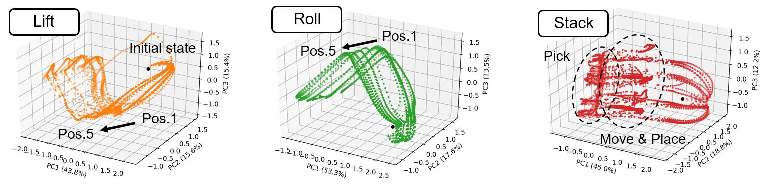}
\caption{
Visualized transition of internal states of LSTM layer in generated Lift, Roll, and Stack tasks.
}
\label{fig:fig8}
\end{figure*}

\subsection{Dataset}
\subsubsection{Robot Task}

To verify the effectiveness of the proposed method, we conducted learning experiments on the cube manipulation task based on language instructions. 
We used Robosuite~\cite{robosuite2020}, a dynamics simulator built on the physics engine MuJoCo~\cite{todorov2012mujoco}. 
For the robot model, we used Panda, which has a total of 8 degrees of freedom (7 for the joints and 1 for the opening/closing gripper). 
Three cubes, of colors red, blue, and green, were placed on the table in front of the robot. 
The robot then performed three tasks: ``Lift a cube,'' ``Stack a cube on another one,'' and ``Roll a cube.'' 
Examples of each task and the instruction sentences are shown in Fig.~\ref{fig:fig4}. 
The cubes were placed at any of the training positions on the table during training dataset collection. 
We obtained 10 patterns for the Lift and Roll tasks and 20 patterns for the Stack task by changing the arrangement of the cubes. 
During data collection, we controlled the robot's posture using a 3D mouse, and the joint angles and camera image sequences were recorded at 10 Hz. 
The average data lengths of the tasks were 61 steps for Lift, 75 steps for Roll, and 133 steps for Stack.

\subsubsection{Training}

We trained the model using the collected motion sequences. 
The input value to the model was normalized to be in the range [0.1, 0.9] and was augmented with Gaussian noise. 
The instruction sentences for each task were prepared as ``Lift the [color] cube'' for the Lift task, ``Roll the [color] cube'' for the Roll task, and ``Stack the [color] cube to the [color]'' for the Stack task. 
In these sentences, the word corresponding to the color of the cube was inserted in place of [color]. 
In the RWKV training, the input sentences were augmented by randomly paraphrasing them into a synonymous sentence using ChatGPT. 
The SLV had a total of 30 patterns, and each was trained for individual sequences.

\subsection{Evaluation Metrics}

During the test phase, we evaluated the proposed method from two viewpoints: position generalization ability and language generalization ability. 
To evaluate the former, the cube was placed at random positions based on the test position range shown in Fig.~\ref{fig:fig4}. 
To evaluate the latter, we used unseen paraphrased sentences (Fig.~\ref{fig:fig4}). 
Specific examples of instruction sentences in the test case will be introduced in the next section.

%% file: sections/5_results.tex
\begin{table}[t]
    \centering
    \begin{tabular}{c|c|cc}
        \multicolumn{4}{c}{TABLE II: Success Rates in Generated Robot Tasks} \\
        \hline
              & & Case 1 (Training position)  & Case 2 (Test position) \\
        \hline \hline
        \multirow{2}{*}{Lift}
        & no ER & 25\% (5/20)   & 25\% (5/20) \\ 
        & w/ ER & 100\% (20/20) & 90\% (18/20) \\ 
        \hline
        \multirow{2}{*}{Roll}
        & no ER & 20\% (4/20)  & 15\% (3/20) \\ 
        & w/ ER & 95\% (19/20) & 90\% (18/20) \\ 
        \hline
        \multirow{2}{*}{Stack}
        & no ER & 5\% (1/20)   & 0\% (0/20) \\ 
        & w/ ER & 85\% (17/20) & 70\% (14/20) \\ 
        \hline
    \end{tabular}
\end{table}

\begin{figure}[t]
\centering
\includegraphics[width=\columnwidth]{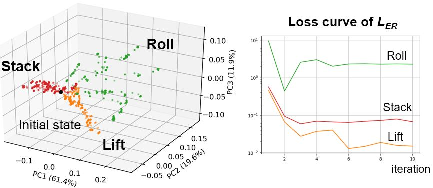}
\caption{
Visualized transition of shared latent variables by PCA, and loss curve of $L_{ER}$.
}
\label{fig:fig9}
\end{figure}

\subsection{Task Performance}

Table II shows the success rates in the generated tasks with and without error regression (ER). 
Three colored cubes were placed at the training position in case 1 or the test position in case 2. 
Note that although the object positions of case 1 have been included in the training dataset, the arrangement and color combinations were not included. 
The results showed that the proposed method performed high success rates in all Lift, Roll, and Stack tasks at the training positions. 
The proposed method also performed high success rates at the test positions, even though slightly lower. 
Examples of the generated tasks for case 2 are shown in Fig.~\ref{fig:fig5}. 
The red cross marks in the figure indicate attention points predicted by the LSTM. 
In the Lift and Roll tasks, the model predicted attention points only for the target cube. 
On the other hand, in the Stack task, the model predicted attention points for each of the two cubes to be operated. 
This allowed the robot to perform tasks according to the position of the object.


When the tasks described earlier were generated, paraphrases based on the instruction sentences shown in Fig.~\ref{fig:fig4} were used. 
Examples of paraphrases included patterns in which verbs change, nouns and adjectives change, and adverbs change (shown in Fig.~\ref{fig:fig5}). 
By estimating a common task expression through the ER phase, the proposed method could accept language instructions in various expressions. 
On the other hand, failure examples included cases where MobileSAMv2 detection failed and cases where the wrong task was selected in the shared latent space. 
These cases can be expected to be improved by changing the LLM or the prompt for the model.

\subsection{Analysis of Model}
\subsubsection{Internal State of RNN}

Next, we verified the behavior of the proposed method by analyzing its internal state. 
Fig.~\ref{fig:fig8} shows the results of principal component analysis (PCA) visualization up to the third principal component in the hidden state of the third layer of the LSTM. 
Each subfigure shows the internal state transition in the corresponding generated task sequence outlined in the previous subsection. 
For all tasks, the internal state of the LSTM corresponds to the cube positions, indicating that the proposed method generalizes the position information of the object. 
Additionally, in the Stack task, the PCA space for picking objects and moving and placing objects are separate. 
This indicates that the proposed method embedded the approach trajectories to the object position into its internal state, and changed this prediction by changing the SLV.

\subsubsection{Shared Latent Space}

The results of visualizing the SLV transition using PCA are shown on the left of Fig.~\ref{fig:fig9}, and the loss curve of $L_{ER}$ for each task is shown on the right of 
Fig.~\ref{fig:fig9}. 
The black point in the figure indicates the initial state (zero vector), whereas the colored points represent the updated states in the iterations during the ER. 
It can be observed that for each task, the SLVs cluster together as the loss function decreases. 
This indicates that the proposed method has an embedded representation suitable for each task in the SLV, and is updated to an appropriate value by the ER of the predicted attention points and language. 
This language generalization ability allows the proposed method to accept diverse language instructions. 
The large variation in the SLVs for the Roll task is due to the wide range of expressions for the instruction sentences. 
For example, unnatural words such as ``spin" and ``twist" were used in the instructions for cube manipulation. 
The aforementioned is a future problem, and it may be solved by introducing uncertainty prediction~\cite{Ren2023a}.

\subsubsection{Self-Attention of Transformer Encoder}

\begin{figure}[t]
\centering
\includegraphics[width=0.9\columnwidth]{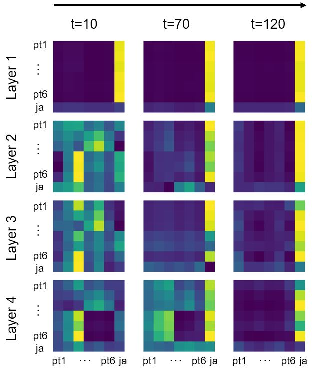}
\caption{
Visualized self-attention layer of Transformer encoder.
}
\label{fig:fig10}
\end{figure}

Finally, by visualizing the self-attention layer of the Transformer encoder, we check how the relationship between the attention points and joint angles was learned. 
Fig.~\ref{fig:fig10} shows the activity of self-attention layers 1--4 in the Stack task shown in Fig.~\ref{fig:fig5}. 
The ja columns and rows in the figure indicate joint angles, pt1--pt3 columns and rows the output of the attention point encoder (i-b), and pt4--pt6 columns and rows another attention point encoder (i-c). 
In layer 1, attention is strongly expressed between the joint angles and each attention point, but as we move to higher layers, the influence of the attention points becomes stronger. 
Based on the time series, as the task approaches its end, the attention points gradually become weaker, and eventually, the joint angle information becomes the most important. 
This result shows that the transformer encoder layers dynamically change the representation as the values of the attention points and joint angles change. 
It also shows that the recurrent unit can handle cases wherein the extracted attention points are redundant.

%% file: sections/6_conclusion.tex

In this paper, we proposed an integrated learning method to efficiently connect LLM and robot-motion learning. 
The proposed method adopted SATrRNN, which predicts attention points for motion prediction, and RWKV as the LLM. 
We then connected the models using shared latent variables. 
During the process of robot-motion generation, an appropriate SLV value is determined based on backpropagation with the prediction errors from the predicted attention points and generated sentences. 
To evaluate the proposed method, we conducted a learning experiment on a simulator for three tasks: Lift, Roll, and Stack, on cubes of three colors. 
Our experimental results showed that the proposed method was able to generate appropriate motions through the error regression phase. 
Analysis of the internal state of the model showed the effectiveness of our method from two perspectives: position generalization and language instruction generalization. 
We plan to extend the proposed method to more complex tasks~\cite{Fujii2022sii}\cite{Saito2023a} and to integrate with LLM task planning~\cite{Hori2024a}.